\documentclass{article}

\usepackage{microtype}
\usepackage{graphicx}
\usepackage{subfigure}
\usepackage{booktabs} 
\usepackage{algorithm}
\usepackage{amsmath}
\usepackage[inline]{enumitem}
\usepackage{amsmath,graphicx,amsfonts,xcolor,url,booktabs,comment,siunitx}

\usepackage{hyperref}

\newcommand{\defeq}{\stackrel{\text{def}}{=}}

\DeclareMathOperator{\Tr}{Tr}
\newcommand{\norm}[1]{\left\lVert#1\right\rVert}

\usepackage[accepted]{icml2021}

\icmltitlerunning{DGA for Federated Domain Adaptation}

\begin{document}

\twocolumn[
\icmltitle{Dynamic Gradient Aggregation for Federated Domain Adaptation}

\icmlsetsymbol{equal}{*}

\begin{icmlauthorlist}
\icmlauthor{Dimitrios Dimitriadis}{MS}
\icmlauthor{Kenichi Kumatani}{MS}
\icmlauthor{Robert Gmyr}{MS}
\icmlauthor{Yashesh Gaur}{MS}
\icmlauthor{Sefik Emre Eskimez}{MS}
\end{icmlauthorlist}

\icmlaffiliation{MS}{Microsoft, Redmond, WA, USA}
\icmlcorrespondingauthor{Dimitrios Dimitriadis}{didimit@microsoft.com}

\icmlkeywords{federated learning, unsupervised training, speech recognition, sequence-to-sequence modeling, optimization}

\vskip 0.3in
]

\printAffiliationsAndNotice{}  

\begin{abstract}
In this paper, a new learning algorithm for Federated Learning (FL) is introduced. The proposed scheme is based on a weighted gradient aggregation using two-step optimization to offer a flexible training pipeline. Herein, two different flavors of the aggregation method are presented, leading to an order of magnitude improvement in convergence speed compared to other distributed or FL training algorithms like BMUF and FedAvg. Further, the  aggregation algorithm acts as a regularizer of the gradient quality. We investigate the effect of our FL algorithm in supervised and unsupervised Speech Recognition (SR) scenarios. The experimental validation is performed based on three tasks: first, the LibriSpeech task showing a speed-up of $7\times$ and $6\%$ word error rate reduction (WERR) compared to the baseline results. The second task is based on session adaptation providing $20\%$ WERR over a powerful LAS model. Finally, our unsupervised pipeline is applied to the conversational SR task. The proposed FL system outperforms the baseline systems in both convergence speed and overall model performance. 
\end{abstract}

\section{Introduction -- Prior Work}
\label{sec:intro}

Distributed Training (DT) is drawing much attention with the goal of scaling the model training processes. Since the training datasets become ever larger, the need for training parallelization becomes more pressing. Different approaches have been proposed over the years~\cite{BNH19}, aiming at more efficient training, either in the form of training platforms such as ``Horovod''~\cite{SeBa18, Abadi+16} or algorithmic improvements like ``Blockwise Model-Update Filtering'' (BMUF)~\cite{chen2016scalable}. These techniques are evaluated on metrics such as data throughput (without compromising accuracy), model and/or training dataset size, and GPU utilization. However, a few underlying assumptions are implied as part of such DT scenarios, i.e., data and device uniformity and efficient network communication between the working nodes. 

Recently, new constraints in data management are emerging. Some of these constraints are driven by the need for privacy compliance of personal data and information~\cite{GDPR}. 
To this end, the Federated Learning (FL) paradigm has been proposed, addressing privacy concerns while still processing such inaccessible data. The proposed approach aims at training ML models, e.g., deep neural networks, on data found on multiples of local worker nodes without the need to exchange any data between the ``coordinator'' and these remote nodes. The general principle is based on training different versions of the model on local data samples while exchanging only updates of the model parameters, such as the network parameters or the corresponding gradients. An additional step of synchronizing these local models and updating the global model at an appropriate frequency is now required. More details about general FL techniques can be found in~\cite{Li+20}. FL is mostly focused on communication efficiency, better optimization~\cite{SSZ13} and/or privacy aspects. 
There are different approaches for FL using either a central server,~\cite{PaY09}, i.e., a ``coordinator or orchestrator,'' or employing peer-to-peer learning, without using a central server,~\cite{LJSC20}. Herein, a single server is responsible for the sampling and communication between  clients, updating models, and adjusting  learning rates. 

A major difference between FL and DT lies on the assumptions made about the properties of the local data sets~\cite{KMR15}.  Since the DT focus is the training of a single model on multiple nodes, a common underlying constraint is that all local data subsets need to be as homogeneous as possible, i.e., uniformly distributed and roughly about the same size. However, none of these constraints are necessary for FL; instead, the data sets are typically heterogeneous, and their size may span several orders of magnitude. It is possible that the data found in each of the clients can be skewed towards different distributions -- especially in SR applications, where accented speech, background noise, or other factors can have an adverse outcome. 

A massively distributed  FL training for Automatic Speech Recognition (ASR) models, like the one herein presented, has not been applied before -- albeit, some work exists for KWS~\cite{Leroy+19}. The e2e models have gained in popularity in ASR tasks because acoustic, language, and pronunciation models of a conventional ASR system can be combined into a single neural network~\cite{Chiu+18}, including ``Recurrent Neural Network Transducer'' (RNN-T)~\cite{Graves12}, ``Listen, Attend and Spell'' (LAS)~\cite{CJLV15} and others.  Herein, the LAS architecture is adopted because it consistently provides the best offline results in our internal test sets. This model includes an encoder, an attention layer, and a decoder. Training of such all-neural models is much simpler than training conventional SR systems, and as such,  easier to federate.

The contributions of the paper are:
\begin{enumerate*}[label=\roman*.]
\item Optimization algorithm: The proposed hierarchical optimization scheme significantly speeds up convergence speed and improves overall classification performance. The ``Generalized FedAvg,''~\cite{Reddi+20} and ``BMUF,''~\cite{chen2016scalable} algorithms overlap with the proposed method.
\item Dynamic Gradient Aggregation: A novel algorithm for de-emphasizing batches of ``bad'' data is presented. Similar algorithms have been investigating in~\cite{Alain+16,BTPG16} for the case of ``Asynchronous SGD''. Further, a data-driven method is applied to the FL setting.
\item Unsupervised Adaptation: An algorithm for unsupervised adaptation of ASR models as part of the FL platform is presented. Unsupervised training using TTS has been proposed before in~\cite{Rosenberg+19}, but it has not been applied to a session adaptation scenario like this one in FL, and 
\item Novelty of the task: The FL approach for ASR model training is investigated for the first time.
\end{enumerate*}

\section{Federated Transfer Learning (FTL) Platform}
\label{sec:proposal}
%

The proposed FL system,~\cite{Dimitriadis+20}, called  ``Federated Transfer Learning'' (FTL) platform, consists of a pool of $K$ clients with a fixed dataset per client. 
Every iteration $t$ consists of processing randomly sampled $N \ll K$  clients and returning them to the pool -- random sampling with replacement. 
Additionally, limiting the processing to the $N$ nodes decreases the latency between iterations and enhances the robustness against rogue nodes or attacks.
Mini-batch optimization methods, extending classical stochastic methods to process multiple data points in parallel, have emerged as a popular paradigm for FL~\cite{LSTS18}. Approaches like  ``Federated Averaging'' (FedAvg)~\cite{KMR15}, a method based on averaging local models after stochastic gradient descent (SGD) updates, are often considered as the golden-standard approach. FedAvg is shown to generalize well while  improving performance in terms of speed-ups. Lately, ``Generalized FedAvg'' was presented in~\cite{Reddi+20, Leroy+19},  with similarities to the proposed hierarchical optimization method.

\subsection{Hierarchical Optimization}

Herein, a different approach for the hierarchical optimization process is proposed. The training process consists of two optimization steps: first, on the client-side using a ``local'' optimizer, and then on the server-side with a ``global'' optimizer utilizing the aggregated gradient estimates. The two-level optimization approach provides both speed-ups due to the second optimizer on the server-side and improved model performance. Scaling in the number of clients becomes straightforward by adjusting the server-side optimizer, as well. The proposed algorithm, i.e., Alg.~\ref{alg:Hier_Optim}, is shown to converge to a better optimum, faster than the centralized training method implemented on Horovod or  BMUF, after data volume normalization~\footnote{The comparison is  in number of iterations till convergence}.

In more detail, the $j^{th}$ client update runs $t$ iterations with $t\in[0, T_j]$, locally updating the seed model $w^{(s)}_T$ (herein shown using the SDG optimizer, without loss of generality) with a learning rate of $\eta_j$,
\begin{equation}
    w^{(j)}_{t+1}=w^{(j)}_t - \eta_j \nabla  w^{(j)}_t
\label{eq:local_update}
\vspace{-0.12cm}
\end{equation}
where $t$ is the local iterations on $j^{th}$-client, i.e., the client time steps, and  $w_t^{(j)}$ the local model and $w_0^{(j)}\defeq w^{(s)}_T$.

The $j^{th}$ client returns a smooth approximation of the local gradient $\tilde{g}_T^{(j)}$ (over the $T_j$ local iterations and $T$ is the iteration ``time'' on the server side) as the difference between the latest, updated local model $w_{T_j}^{(j)}$ and the previous global model $w^{(s)}_T$ 
\begin{equation}
   \tilde{g}_T^{(j)} = w^{(j)}_{T_j} -   w_T^{(s)}
    \label{eq:local_gradient}
    \vspace{-0.12cm}
\end{equation}
Since, estimating the gradients $g_T$ is extremely difficult, hereafter the approximation $\tilde{g}_T^{(j)}$ is used instead.

The gradient samples $\tilde{g}_T^{(j)}$ are weighted and aggregated, as described in Section~\ref{sec:aggregate}
\begin{equation}
   g^{(s)}_T =  \sum_j{\alpha_T^{(j)} \tilde{g}_T^{(j)}}
\label{eq:global_grad}
 \vspace{-0.12cm}
\end{equation}
where $\alpha_T^{(j)}$ are the aggregation weights.

The global model $w^{(s)}_{T+1}$ is updated as in~(\ref{eq:global_grad}) (here also shown using SGD, although not necessary), 
\begin{equation}
   w^{(s)}_{T+1} = w_T^{(s)} - \eta_s g^{(s)}_T.
    \label{eq:global_update}
    \vspace{-0.12cm}
\end{equation}

The process described in Equation~\ref{eq:server_training} is a form of ``\emph{Online Training}''~\cite{Parisi+18, SPLH17}.  While updating  the seed model drifts further from the original task. In order to mitigate this drifting,  an additional training step is proposed (with held-out data matching  the tasks in question) on the server-side.
The model updates are regularized in a direction matching the held-out data. As such,  the model does not diverge too much from the task of interest. 
\begin{equation}
    w^{(s)}_{\tilde{t}+1} = w^{(s)}_{\tilde{t}} - \eta_w \nabla  w_{\tilde{t}}
    \label{eq:server_training}
\end{equation}
This training step on the server-side can be seen as an example of ``\emph{Naive Rehearsal}''~\cite{HLRK18, Parisi+18}, replaying previously seen data.

\begin{algorithm}[htb]
\caption{Hierarchical Optimization}
\label{alg:Hier_Optim}
\begin{algorithmic}[1]
    \STATE {\bfseries Input: }( $w_0^{(s)},\ \bf{x}^{(j)}_T$)
    \WHILE{Model $w_T^{(s)}$ has not converged}
        \FOR{$j\ in\ [0,N]$}
            \STATE Send seed model $w_T^{(s)}$ to $j^{(th)}$ clients 
            \STATE Train local models $w^{(j)}$ in $j^{th}$-node with data $x_T^{(j)}$
            \STATE Estimate a smooth approximation of the local gradients $\tilde{g}_t^{(j)}$ for $j^{th}$-node
            \STATE $j \leftarrow j+1$
        \ENDFOR
        \STATE Estimate weights $\alpha^{(j)}_T$ per DGA algorithm
        \STATE Aggregate weighted sum of gradients, Equation~\ref{eq:global_update}
        \STATE Update global model $w_T^{(s)}$ using aggregated gradient 
        \STATE Update global modal on held-out data, Equation~\ref{eq:server_training}
    \ENDWHILE
\end{algorithmic}
\end{algorithm}
The convergence speed of training due to the hierarchical optimization scheme is improved by a factor of $2\times$, without any negative impact on performance. Also, the communication overhead is significantly lower since the models are transferred twice per client and iteration (instead of continuously transmitting the gradients, as in FedSGD.

\subsection{Unsupervised Training}

Accurate labels are not always available in many FL scenarios and/or ASR tasks. 
In this work, we present two unsupervised training methods, either utilizing multiple hypotheses~\cite{KDGG20} or based on available local text. The first algorithm processes the $N$-best hypothesis of the speech recognizer as a sequence of soft labels. These $N$-best hypotheses, even with $N$ relatively small, have coverage of around $90\%$ of the correct labels. The network is updated with these soft labels as part of multi-task training, where the loss of each task is weighted based on the DGA, as in Section~\ref{sec:aggregate}. By doing so, we can alleviate degradation caused by hard labels, as in the other adaptation methods~\cite{HWG16}. 
The second approach is to adapt the model with TTS-based audio hierarchically. A mix of audio from TTS and randomly sampled speech is used as input for this approach. First, the LAS model is adapted with organization-level relevant data, and then, it is adapted with session-specific data in a federated manner. The first step creates a new seed model with the decoder lightly matching the expected session-based data. This seed model is used as the starting point for running the FTL pipeline described above on TTS-data per session. The TTS-based data causes the LAS model to diverge from the original one significantly, so the randomly sampled `real' data is used to regularize the training process, as in Equation~\ref{eq:server_training}. 

\section{Dynamic Gradient Aggregation}
\label{sec:aggregate}

Training with heterogeneous local data poses additional challenges, especially for the aggregation step, Equation~\ref{eq:global_update}. 
%
As a result of such heterogeneous distributions, the gradients can point to different directions and the aggregation process  becomes noisier due to this diversification. 
Gradients based on local training, where losses are of similar magnitude, are expected to move the model similarly; thus, the aggregated gradients are  expected somewhat aligned. Such alignment of the aggregated gradients is shown to be beneficial both for the convergence speed and performance, per the experimental section below. 

Gradients that deviate from the rest should be processed differently: the proposed approach uses weights during the aggregation step, i.e., the contribution of some components is de-emphasized by weighting the local gradients, $\tilde{g}_T^{(j)}$, in~(\ref{eq:local_gradient}). The proposed algorithm is called ``\emph{Dynamic Gradient Aggregation}'' (DGA). Two flavors of DGA  are herein proposed: first, the  $\text{DGA}_{SM}$ using the training losses as  weights, and the ``data-driven'' approach, where a neural network is trained to infer the weights.
In some tasks,  DGA does not significantly affect the overall WER performance (e.g., in the LibriSpeech task, where data is more homogeneous); however, it makes the training convergence significantly faster. 
On the other hand, the DGA  approach  significantly affects the convergence speed and classification performance in other tasks, e.g., unsupervised training where the label quality varies significantly or in diverse local data distributions.  
The weighting process can be seen as a type of regularization, decreasing the variance of the aggregated gradients. The analysis can be found in the Appendix~\ref{app:dga_theory}.

The  $\text{DGA}_{SM}$ algorithm  utilizes the negative training loss coefficients $\mathcal{L}$, as weights $\alpha^{(j)}_t \sim \mathcal{L}(\cdot)$. These weights are normalized when passed through a $\text{Softmax}(\cdot)$ layer,
\begin{equation}
\label{eq:softmax_weights}
    \alpha^{(j)}_T=\exp(\mathcal{-\beta L}^{(j)}_T)/\sum_i{\exp{(-\beta \mathcal{L}^{(i)}_T})}
\end{equation}
where $\beta$ is the temperature of the Softmax function. 
According to Equation~\ref{eq:softmax_weights}, the aggregation weights are smaller for nodes with larger values of the corresponding losses.

The $2^{nd}$ approach is based on Reinforcement Learning. The weights are inferred by a network, trained with RL rewards.  Herein, this approach is called  ``RL Weighting'' (or $\text{DGA}_{RL}$), Algorithm~\ref{alg:RL_Training}.
An agent perceives an ``observation'' from the environment. The rewards  depend on how good the agent's action is, based on a specific, predefined reward policy. This agent takes action in order to optimize the interaction with the environment according to such rewards policy while inducing new states to the system. Then, updated observations and a new reward are acquired based on such a new state. Our approach is based on Active  Reinforcement Learning, where  rewards depend  on the action selected~\cite{EVD08}.  

A neural network $RL(\cdot)$ is trained with Reinforcement Learning to infer the weights, based on the input features $\bf{x}_T^{(j)}$ from each of the $j$-clients. The agent decides on the values of the gradient weights aiming at improving the overall CER performance of the model by learning an optimal weighting strategy. The input features consist of the training loss coefficients and the gradient statistics.

\begin{algorithm}[htb]
\caption{Dynamic Gradient Aggregation Based on Reinforcement Learning}
\label{alg:RL_Training}
\begin{algorithmic}[1]
   \STATE {\bfseries Input:} $\tilde{g}^{(j)}_T, \bf{x}^{(j)}_T$
    \STATE $RL(\cdot)$ Model Initialization
    \WHILE{Model $w_T^{(s)}$ has not converged}
        \STATE Read the Observations $\bf{x}^{(j)}_T$
        \STATE Estimate Weights $a^{(j)}_T \leftarrow  RL(\bf{x}^{(j)}_T) $
        \STATE Estimate Weights $b^{(j)}_T \leftarrow \text{Softmax}(\bf{x}^{(j)}_T)$, from Equation~\ref{eq:softmax_weights}
    
        \STATE Estimate $\tilde{W}^{(s),a}_T$ and $\tilde{W}^{(s),b}_T$ based on two sets of aggregation weights $a^{(j)}_T, \  b^{(j)}_T$ with Equation~\ref{eq:global_update}
    
        \STATE Estimate $\text{CER}_{a, T},\ \text{CER}_{b, T}$ for the new models $\tilde{W}^{(s),a}_T, \tilde{W}^{(s),b}_T$     
        
        \IF {$\text{CER}_{a, T}- \text{CER}_{b, T}>\theta$}
            \STATE $w^{(s)}_{T+1} \leftarrow \tilde{w}^{(s),a}_T$
            \STATE $r_T=R$
        \ELSIF{ $ \text{abs}(\text{CER}_{a, T}- \text{CER}_{b, T}) \leq \theta$ }
            \STATE $w^{(s)}_{T+1} \leftarrow \tilde{w}^{(s),a}_T$
            \STATE $r_T=0.1R$
        \ELSE
            \STATE $w^{(s)}_{T+1} \leftarrow \tilde{w}^{(s),b}_T$
            \STATE $r_T=-R$
        \ENDIF
        \STATE Update model $RL(\cdot)$ on reward $r_T$ 
    \ENDWHILE
\end{algorithmic}
\end{algorithm}

The ``actions'', i.e., the weights, are predicted from the results they incur, i.e., the ``Character Error Rate'' (CER) on the validation set in every time step $T$. 
This reward policy is based on the CER performance, i.e., the ``environment,'' the ``action,'' and ``state'' $s_{T+1}$ is the new aggregated model.
In more detail, the ``environment'' in every iteration $T$ is the gradient components $\tilde{g}_T^{(j)}$, the states $s_T$ are described by the input features $\bf{x}^{(j)}_T$, and the action vector $\alpha^{(j)}_T$ is the aggregation weights. The policy is dictated by the output CER performance (on the validation set), 
as  in Alg.~\ref{alg:RL_Training}.
The reward policy is based on the CER performance of two different networks that are trained with the aggregated gradients~(\ref{eq:local_gradient}). These two networks are versions of the  same seed model, after training with either the inferred weights or the Softmax-based ones, Equation~\ref{eq:global_update}. Depending on the comparative results, a reward $r_T$ is provided, and the new state $s_{T+1}$ is estimated. The threshold and reward parameters $\theta,\ R$ are part of the reward policy as detailed in Alg.~\ref{alg:RL_Training}.
Herein, the input features $\bf{x}_T^{(j)}$ are the combination of the training loss coefficients (as described above) augmented by the gradient magnitude mean and variance values. The  statistics are estimated on the gradients $\tilde{g}_T^{(j)}$, Equation~\ref{eq:local_gradient}.  
%


\section{Experiments and Results}
\label{sec:experiments}

Three datasets are used as the experimental test-beds, the LibriSpeech task~\cite{PCPK15} (LS task) for supervised training,  an internal dataset based on Powerpoint presentations and meeting transcriptions (CTS) for the unsupervised tasks.  A LAS model is used~\cite{CJLV15}, consisting of a 6-layer bLSTM, with dropouts for the encoder, 2 layers of uni-directional  LSTM  for the decoder, and a conventional location-aware content-based attention layer with a single head. The input features are 80-dim log mel filter-bank energies, extracted every $10\ msec$ with 3 consecutive frames stacked, and $16k$  subwords based on a  unigram language model are used as the recognition unit. For the $1^{st}$ scenario, the baseline model is a state-of-the-art LAS model trained using Horovod on the entire training set of $920h$. This model's performance provides the lower bound of the WER for the particular ASR task since all the data is used in a centralized manner. For the $2^{nd}$ scenario, a LAS model of similar architecture, trained on $75k$ hours  of speech, is used as the seed model.

\subsection{Supervised Training FTL Experiments}
\label{subsec:ftl_experiments}

For the supervised experiments, the dataset contains about $1k$ hours of speech from $2.5k$ speakers reading books. The training set is split into two parts of 460h each, with no overlapping speakers. The first part is used to train a seed model. The $2^{nd}$ part of the dataset is used to simulate online training under the FL conditions. Two different FL scenarios are examined: first, i.i.d.\ split of the training set in 7 randomly split parts, second, a non-i.i.d.\ segregation based on the $1100$ speaker labels. 
The number of randomly sampled clients $N$ in our experiments varied from 25 to 400, with higher $N$ being better.
We set $N=100$ as a compromise between communication overhead and memory.

The weighting approach for $\alpha_T^{{j}}$ is either based on the training loss ($\text{DGA}_{SM}$) or inferred by the $RL(\cdot)$ network ($\text{DGA}_{RL}$). The $RL$ network is a 5-layer DNN with ReLU activations and a bottleneck layer ($2^{nd}$ layer to last). The input layer size is $3N$, and the output layer $N$. The reward policy is $+1$ if the weights provide a better CER value compared to the loss-based weights, $-1$ in the opposite case, and $+0.1$ when the performance of the two cases is similar. The network has a memory of the previous $1k$ instances, sampling a mini-batch of $32$ instances  per  training iteration.
\begin{table}[htb]
	\centering
	\begin{tabular}{|c|l|c|}
	    \hline
		\multicolumn{3}{|c|}{LibriSpeech Task}\\ \hline\hline
		                            & Training Scenario & WER (\%)\\ \hline
		Central.         & Offline (lower bound) & 4.00\% \\ \cline{2-3}\cline{2-3}
		                            & Training on 1st 50\% of LS(seed) & 5.66\% \\ \cline{2-3}
		                            & Online   & 4.61\% \\ \hline
		FTL              & FedAvg                        & 4.55\% \\ \cline{2-3}
		                            & Hier.  Optim. (i.i.d.)        & 4.51\% \\ \cline{2-3}
		                            & Hier.  Optim. (non-i.i.d.)    & 4.45\% \\ \cline{2-3}
		                            & \ \   +  $\text{DGA}_{SM}$               & 4.41\% \\ \cline{2-3}\cline{2-3}
		                            & \ \   +  $\text{DGA}_{RL}$               & 4.40\% \\ \hline
	\end{tabular}
	\caption{System evaluation on the LibriSpeech task, for online learning with LAS.}
    \label{tab:LS}
    \vspace{-0.3cm}
\end{table}

The top 3 rows in Table~\ref{tab:LS} are with centralized training, with the lower bound in performance coming from the model trained on the entire dataset (offline training). The $4\%$ WER for this model appears in line with the literature. The $2^{nd}$ model (``online'' row) is based on the seed model initially trained on the 1st half of the data till convergence. The model is online trained with the $2^{nd}$ half of the data (both trained with Horovod). 
Then, the seed model is further refined with FTL  by training on unseen data. Different strategies for model aggregation were investigated, such as model averaging (``FedAvg'' row) or hierarchical optimization using optimizers such as Adam, SGD, etc. For the FedAvg system, the model averaging is performed on the server, while the SGD optimizer is used for training on the client's side. Combinations of the server/client optimizers were also investigated. The differences in performance of these combinations of optimizers were rather limited, and for the sake of space, are not further elaborated here. However, a state-less optimizer on the client-side is expected because the initial model is changing after each iteration, and therefore, keeping the state  of the local optimizer does not make sense. Throughout the paper, the combination of Adam/SGD for the server/client is adopted. 

The non-i.i.d.\ partition of the data is based on the speaker labels, i.e., $1100$ partitions and an equal number of clients. As mentioned above, $100$ clients per iteration are sampled out of the pool of clients. The transition from an i.i.d.\ data split, i.e.,  7 partitions, to non-i.i.d.\ per-speaker partition, did not deteriorate the overall model performance. 

The proposed DGA algorithm addresses the data heterogeneity by de-emphasizing the gradients from clients badly modeled. The FedAvg system requires around 800 iterations for convergence. Such a system is too slow, and henceforth, it will not be considered as the state-of-the-art baseline. Our baseline system is based on hierarchical optimization but without DGA. Even though the overall performance was not impacted (this is only true for the case of the LS task), the overall convergence speed was improved by a factor of  $~1.5\times$ compared to the baseline system and $~7\times$ compared to the FedAvg system. In more detail, approximately 384 iterations are required for the case of unweighted aggregation against only 224 for the $\text{DGA}_{SM}$. In the case of $\text{DGA}_{RL}$, the number of required iterations is even lower, decreased by an additional factor of $1.5\times $, requiring only 144 iterations. The variations in performance between different approaches are limited; the task is quite homogeneous. However, improvements in the other tasks, e.g., adaptation on presentation sessions, are presented.  

\begin{figure}[tb]
  \centering
  \includegraphics[width=1.0\columnwidth]{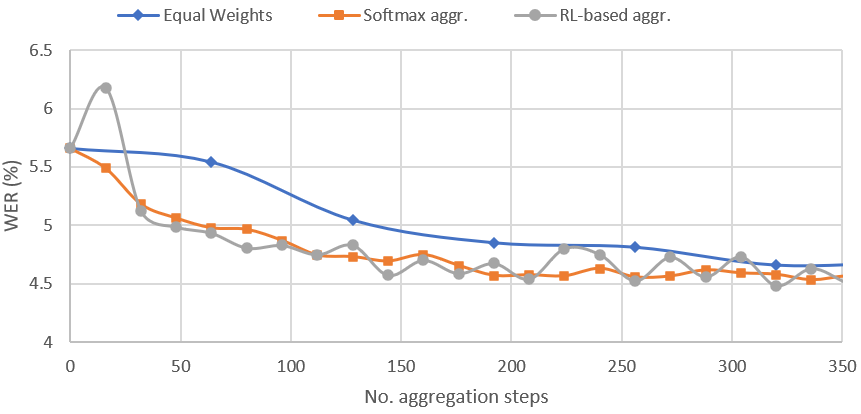}
  \caption{Training on LibriSpeech corpus for 3 weighting scenarios, i.e., uniform weights, $\text{DGA}_{SM}$ and $\text{DGA}_{RL}$ weights.}
  \label{fig:iterations}
  \vspace{-0.38cm}
\end{figure}
The convergence plots for the different weighting scenarios, i.e., uniform weights, $\text{DGA}_{SM}$ and $\text{DGA}_{RL}$, for the LS task, are shown in Figure~\ref{fig:iterations}. 
The RL-based aggregation curve shows ripples (particularly wide at the beginning of the training process), but this is expected since the RL network starts from a random state and convergences later on. Further, the performance of the RL- and Softmax-based weighting schemes seem to converge after a few iterations. This is also expected since the rewards policy is based on the Softmax- performance. The RL network learns Softmax-based behavior after a while. However, it is consistently outperforming that system in terms of convergence speed. The LS dataset is quite homogeneous in terms of audio quality, so the proposed algorithm is not expected to perform vastly differently in terms of WER performance. However, the RL-based system outperforms the uniform-weighted one in other tasks with improvements up to $11\%$ WERR.

\subsection{Hierarchical Unsupervised Training with Federated Session Adaptation}
\label{subsec:unsupervised_experiments}

In the case of unsupervised training, our approach is to adapt the LAS models hierarchically: first, adapt the seed model to organizational-level tenant data, and then perform session adaptation based on the FTL platform. The dataset used here consists of TTS data and a random sample of real speech already used for training the LAS model; the use of real data prevents the model from overfitting to the TTS data. The TTS corpus is $3.5k$ hours of audio, based on $2.8M$ written sentences, i.e., $32M$ words with a dictionary size of $172k$ unique words, of internal documents and emails from Microsoft employees. This written corpus is pushed through the Neural TTS Microsoft service. The test set is based on 4 presentations of about $40min$ each, with 1230 sentences, $12k$ words, and a vocabulary of $2.5k$ words. The text data found in the presentation slides were also used to synthesize the speech data (using the TTS engine as described below) with a total size of $1.3h$. The initial LAS model is trained on $75k$ hours of speech. This seed model is adapted with the TTS data generated from the tenant text data in a centralized way. Several scenarios for this adaptation step are investigated, performing subspace adaptation, i.e., changing the encoder, decoder, or both, and including different sources of the TTS speech data: the tenant TTS data only, tenant and presentation slide TTS data, and the mixture of the TTS and real-speech data. Now, the final adapted model matches closer the content of the presentation slides.

The adapted-to-the-tenant-text-data model is now used as the starting model for the second adaptation step. This $2^{nd}$ adaptation step is based on the FTL platform, where the 4 meetings and the new model are used for the final inference step. 
The new model is adapted iteratively to the presentation-related text using the TTS-based audio. The DGA step is also applied to weight the input gradients. However, we noticed that the model overfits very fast on the synthetic TTS data, with the overall performance steeply deteriorating. To address this issue, we have added real-speech data on the server-side training to regularize the process. The real-speech audio is randomly picked from the training set. Note here that this random subset of speech has already been used for training the initial LAS model -- no need for held-out data. The addition of this set reduces the model drift significantly while improving the overall recognition performance. This step mitigates any impact from Catastrophic Forgetting.
\begin{table}[t]
	\centering
	\begin{tabular}{|c|l|c|c|}
	    \hline
		\multicolumn{4}{|c|}{Presentation-based Session Adaptation}\\ \hline\hline
		                                &Train. Scenario          & Adapt. Comp.  & WER \\ \hline
		Central.            & Baseline                & None          & 6.86\% \\ \cline{2-4}\cline{2-4}
		                                & Tenant Text  & Encoder   & 8.71\% \\ \cline{3-4}
		                                                    && Decoder  & 6.41\% \\ \cline{2-4}
		                                & \  + TTS + DGA$_{RL}$          & Decoder  & 6.46\% \\ \cline{2-4}
		                                & \ \ \ \   + Real speech & Decoder  & 6.29\% \\ \hline\hline
		FTL                             & TTS+Real speech         & Decoder  & 5.51\% \\ \hline\hline
	\end{tabular}
	\caption{Hierarchical Unsupervised Training with Federated Session Adaptation.}
    \label{tab:PPT}
    \vspace{-0.3cm}
\end{table}

The hierarchical approach shows an $~20\%$ WERR improvement over the original model performance. We have investigated TTS adaptation on the speakers' voice (each of the 4 presentations is assumed to contain a single speaker), but no additional benefits in performance were found.

\subsection{Semi-supervised Learning with DGA}
\label{subsec:cts_experiments}
\begin{table}[t]
\begin{center}
\begin{tabular}{|c|c|c|}
  \hline
  Model type & Adapt.   & WER (\%)\\ \hline\hline
  Teacher model (1GB + LM) &       No       &  20.80\% \\ \hline
  Large student model (1GB) & No & 22.50\% \\ 
                                           & Yes  & 20.40\% \\ \hline
  Small student model (350MB) & No & 22.64\% \\
   & \  Yes  & 21.48\% \\ \hline
  
\end{tabular}
\caption{Federated adaptation with $\text{DGA}_{RL}$ via knowledge distillation and self-training.}
\vspace{-0.35cm}
\label{tab:meeting}
\end{center}
\end{table}

To verify the efficacy of  DGA, we ran another set of SR experiments on meeting data in a semi-supervised scenario. The experiments followed the same setup as described in~\cite{Wong2021}. Audio was captured with a microphone array from internal Microsoft meetings with an average of 7 active participants each. Meeting participants interact with each other in an unrestricted way. The meeting audio data, thus, contain conversational overlapping speech as well as background noise; as in~\cite{yoshioka2019meeting}. Each meeting lasted up to 1 hour. 111 meetings were used as a test set. The total amount of the test data is approximately $58$h.  Additional untranscribed speech data form 711 meetings are used for domain adaptation. 

The teacher acoustic model is a hybrid latency-controlled and layer-trajectory bLSTM network~\cite{Sun2019} with 6 layers of 1024 nodes. Two different sizes of LAS networks are used as student models, with size of 350MB and 1GB, respectively. 
The hybrid and LAS models were initially trained on  $75k$ hours of anonymized data from a variety of Microsoft applications. 
For generating the soft labels, the adaptation data were first decoded by using the hybrid teacher model with a 5-gram
language model. The N-best results were then re-ranked with a neural network language model, comprising 2 LSTM layers with 1024 nodes, trained on text
data crawled from the internet and meeting-related scenarios. The gradients are separately computed for each meeting by sequence-level knowledge distillation and self-training~\cite{KDGG20} with the N-best soft-labels. The final model update is done by combining the gradient of each meeting with $\text{DGA}_{RL}$, as described in Section~\ref{sec:proposal}. 

Table~\ref{tab:meeting} shows the WER for both the teacher and student LAS networks adapted with DGA. It is clear from table~\ref{tab:meeting} that our FL framework with DGA can improve recognition accuracy in the semi-supervised learning scenario. It is also worth mentioning that our FL platform can achieve better accuracy on the compressed model.

\section{Discussion and Future Work} 
\label{sec:conclusion}
In this paper,  novel optimization algorithms are presented, addressing challenges unique to the ASR scenarios. This novel approach of weighting the gradients between mini-batches allows for enhanced convergence speed-ups and improved model performance. The proposed gradient aggregation scheme acts as a regularizer, de-emphasizing batches where the data are not well modeled. The FL application on ASR tasks is also presented here. Our weighted gradient aggregation algorithm shows a $7\times$ speed-up and $6\%$ WERR on the LS task,  $20\%$ WERR for a session adaptation task, and improved performance in a model compression task. 

Although the experimental part is focused on the ASR tasks, the algorithms are shown to generalize to other tasks, such as face identification and NLP applications.


\bibliography{main}
\bibliographystyle{icml2021}

\appendix
\section{Dynamic Aggregation Algorithm -- Theoretical Proof}
\label{app:dga_theory}

The gradient aggregation of Equation~\ref{eq:global_grad}, can be seen as (dropping the subscript $T$)
\begin{equation}
    \mathbb{E}_p[g^{(s)}]=\int{p(g) g^{(s)} dg} = \mathbb{E}_\alpha \left[ \frac{p(g)}{\alpha(g)}g^{(s)} \right]
    \simeq \frac{1}{N}\sum_{j=1}^N{\alpha^{(j)}g^{(j)}}    
\end{equation}
where $p(\cdot)$ is the probability density function of the random variable $\mathcal{G}$ and $g^{(s)}$ is the  function mapping gradients to the gradient space $\mathbb{R}^d$, with $d$ the gradient dimension. The mean value of the gradient $g^{(s)}$ can be approximated by the weighted sum of the gradient samples  $g^{(j)}$ for $j=1,\dots,N$ based on the client data, as long as $a^{(j)}~\sim p(g^{(s)})$ and $\alpha^{(j)}\geq0$. Equally, if the training step in the server and clients is considered as a large-batch process then $g^{(s)}\sim\nabla_\theta\mathcal{L}(\cdot)$, as in Section~\ref{sec:aggregate}, and $g^{(j)}$ are samples of the random variable $\mathcal{G}$.

It is shown, that reducing the variance of the gradient $g^{(s)}$ can improve the convergence speed during training.
In a multi-dimensional space, it is shown that an equivalent criterion of variance minimization is  minimizing the eigenvalues of the Covariance matrix $\Sigma(\cdot)$ of the weighted gradients, or equally $\min_\alpha \Tr(\Sigma(\alpha))$, 
where $\Sigma(\alpha)$ is a function of the gradient weights $\alpha=\{\alpha^{(j)}\}_{j=1}^N$, and we will minimize the $\Tr\left(\Sigma(\alpha)\right)$ by optimizing these weights~\cite{Alain+16},  

\begin{equation}
    \Tr(\Sigma(\alpha))
    =\mathbb{E}_\alpha \left[ \norm{\frac{p(g)}{\alpha(g)} \nabla _\theta \mathcal{L}(\cdot)}_2^2\right]-\norm{\mu}_2^2
    \label{eq:orig_sampler}
\end{equation}
where $\mu=\mathbb{E}_p[\nabla _\theta \mathcal{L}(\cdot)]$. Based on Eq.~\ref{eq:orig_sampler}, the optimal weights $\alpha^*$ are given by 
\begin{equation}
    \alpha^*=\frac{1}{Z}p(g)\norm{ \nabla_\theta \mathcal{L}(\cdot)}_2
    \label{eq:optimal_weights}
\end{equation}
where $Z=\int{p(g)\norm{\nabla _\theta\mathcal{L}(\cdot)}_2 dg}$, according to~\cite{Alain+16}. 
\\
\\
\\
\\
Proof.
\begin{equation}
    \begin{split}
        \Tr(\Sigma(\alpha)) &= \Tr\left( \mathbb{E}_\alpha \left[ \left(\frac{p}{\alpha}
         g^{(s)}-\mu\right)\left(\frac{p}{\alpha} g^{(s)}-\mu\right)^T\right] \right) \\
         & =  \Tr\left(  \mathbb{E}_\alpha \left[\left(\frac{p}{\alpha} g^{(s)}\right)\left(\frac{p}{\alpha} g^{(s)}\right)^T\right]-\mu\mu^T\right) \\
         & =  \mathbb{E}_\alpha \left[  \Tr\left( \left(\frac{p}{\alpha} g^{(s)}\right)\left(\frac{p}{\alpha} g^{(s)}\right)^T\right)\right]-\norm{\mu}_2^2 \\
         & =  \mathbb{E}_\alpha \left[  \norm{\frac{p}{\alpha} g^{(s)}}_2^2\right]-\norm{\mu}_2^2
    \end{split}
 \label{eq:theory_analysis}
 \end{equation}
 
 From Jensen's inequality, we get,
\begin{equation}
    \begin{split}
        \mathbb{E}_\alpha \left[  \norm{\frac{p}{\alpha} g^{(s)}}_2^2\right] &\geq  \left(  \mathbb{E}_\alpha \left[\norm{(\frac{p}{\alpha} g^{(s)}}_2\right]\right)^2 \\
         &=   \left(  \int{\frac{p}{\alpha} \norm{g^{(s)}}_2 dg}  \right)^2 =  \left(  \mathbb{E}_p\left[  \norm{g^{(s)}}_2 \right]  \right)^2 
    \end{split}
 \label{eq:theory_ineq}
 \end{equation}
 when $\alpha, p \geq 0$. Based on Equation~\ref{eq:theory_ineq}, the lower bound for $\Tr(\Sigma(\alpha))$ is $\left(  \mathop{\mathbb{E}}_p\left[  \norm{g^{(s)}}_2 \right]  \right)^2-\norm{\mu}_2$.

 Now, replacing the weights $\alpha$ with the $\alpha^*$ from Equation~\ref{eq:optimal_weights}, we get
 \begin{equation}
    \begin{split}
            \Tr(\Sigma(\alpha^*)) &= \mathbb{E}_{\alpha^*} \left[ \norm{(\frac{p(g)}{\alpha^*} g^{(s)}}_2^2\right]-\norm{\mu}_2^2 \\
            &= \int{\alpha^* \left(\frac{p(g)}{\alpha^*} \right)^2 \norm{g^{(s)}}_2^2 dg}-\norm{\mu}_2^2  \\
            &= \int{ \frac{p^2 Z}{p(g)\norm{ g^{(s)}}_2}  \norm{g^{(s)}}_2^2 dg}-\norm{\mu}_2^2 \\
            &= \left( \mathbb{E}_p\left[  \norm{g^{(s)}}_2 \right]  \right)^2-\norm{\mu}_2
    \end{split}
    \label{eq:theory_optimal_weights}
 \end{equation}
 So, we have proved that $\alpha^*$ provides the optimal solution based on the lower bound estimated in Eq.~\ref{eq:theory_ineq}.

For a function $h:~\mathcal{G} \to \Re^+$, then the Covariance matrix is given by
\begin{equation}
    \Tr(\Sigma(\alpha))=\left( \int p(g)h(g)dg \right)\left( \int p(g)\frac{\norm{\nabla _\theta\mathcal{L}(\cdot)}_2^2}{h(g)}dg \right)   -\norm{\mu}_2^2
    \label{eq:sampler}
\end{equation}

In the case when $p(g^{(s)})$ is not known but we can draw samples, i.e. $g^{(j)}$, from that distribution $p(g^{(s)})$, we have that the weights $\alpha$ are approx. proportional to $p(g^{(s)})h(g^{(s)})$ and
%
\begin{equation*}
    \tilde{\alpha}_j=h(g^{(j)}),\ \forall g^{(j)}\in g^{(s)}
\end{equation*}
and after normalizing to unity, $\alpha_j=\tilde{\alpha}_j/\sum_j \tilde{\alpha}_j$. Based on the sampled weights $\alpha_j$, Eq.~\ref{eq:sampler} becomes,
\begin{equation}
        \Tr(\Sigma(\alpha))= \left( \frac{1}{N^2}\sum_{j=1}^N{\frac{\norm{\nabla _\theta \mathcal{L}^{(j)}}_2^2}{\alpha_j}}\right)
    \label{eq:weighted_sampler}
\end{equation}

If we replace $h(g^{(j)})=\exp(-\beta \norm{g^{(j)}})$, then we get a form of the DGA algorithm as described in Section~\ref{sec:aggregate}.

Finally, the analysis is based on $\norm{g^{(j)}}$ but it's safe to assume that $\norm{g^{(j)}}\sim \mathcal{L}^{(j)}(\cdot)$, where $\mathcal{L}^{(j)}(\cdot)$ is the training loss for the $j^{th}$ client, since the client optimizer is SGD (without momentum). This assumption has been experimentally verified, as well.

\end{document}